\documentclass[conference]{IEEEtran}
\IEEEoverridecommandlockouts

\usepackage{times,latexsym}
\PassOptionsToPackage{breaklinks}{hyperref}
\usepackage{breakurl}
\usepackage[T1]{fontenc}

\usepackage{xspace,mfirstuc,tabulary}

\usepackage{cite}
\usepackage{algorithm}
\usepackage{algorithmic}
\usepackage{graphicx}
\usepackage{textcomp}
\usepackage{color}
\usepackage{xcolor}
\usepackage{array}
\usepackage{pifont}
\usepackage{tabularx}
\usepackage{adjustbox}
\usepackage{booktabs,dcolumn}
\usepackage{makecell}
\usepackage{multirow}
\usepackage{multicol}
\usepackage{tablefootnote}
\usepackage{threeparttable}
\usepackage{float}
\usepackage{subfigure}
\usepackage{paralist}
\usepackage{comment}
\usepackage{enumitem}
\usepackage{enumerate}
\usepackage{mathtools}
\usepackage{ulem}
\usepackage{amsmath,amsthm,amsfonts,amssymb,bm,stmaryrd}
\usepackage{soul}
\usepackage{url}

\usepackage{fmtcount}
\usepackage{flushend}
\usepackage{thmtools}
\usepackage{colortbl}
\definecolor{mygray}{gray}{.9}

\usepackage{balance}

\usepackage{CJKutf8}

\usepackage{tikz}
\usepackage[edges]{forest}
\definecolor{hidden-draw}{RGB}{205, 44, 36}
\definecolor{hidden-blue}{RGB}{194,232,247}
\definecolor{hidden-orange}{RGB}{243,202,120}
\definecolor{hidden-yellow}{RGB}{242,244,193}
\definecolor{hidden-red}{RGB}{255,204,204}
\definecolor{hidden-slm}{RGB}{108,166,205} 
\definecolor{hidden-slm1}{RGB}{176,226,255}
\definecolor{hidden-slm2}{RGB}{135,206,255}
\definecolor{hidden-slm3}{RGB}{99,184,255} 
\definecolor{hidden-llm}{RGB}{124,205,124} 
\definecolor{hidden-llm1}{RGB}{188,238,104}
\definecolor{hidden-llm2}{RGB}{162,205,90}
\definecolor{tree-level-1}{RGB}{245,20,85}
\definecolor{tree-level-2}{RGB}{246,86,118}
\definecolor{tree-level-3}{RGB}{248,177,193}
\definecolor{tree-leaf}{RGB}{176,230,198}
\definecolor{slm}{RGB}{24,116,205}
\definecolor{llm}{RGB}{34,139,34}
\definecolor{finding}{RGB}{205,0,0} 

\definecolor{exemplar1}{RGB}{136,98,148}
\definecolor{exemplar2}{RGB}{148,210,242}
\definecolor{knowledge1}{RGB}{249,219,152}
\definecolor{knowledge2}{RGB}{255,245,220}

\usepackage{pgfplots}
\pgfplotsset{compat=1.14}
\usetikzlibrary{pgfplots.groupplots}

\let\oldbibitem\bibitem
\def\bibitem{\vfill\oldbibitem}

\newcommand{\ourslong}{Information Extraction}
\newcommand{\ours}{IE}

\newcommand{\eg}{\hbox{\emph{e.g.}}\xspace}
\newcommand{\etc}{\hbox{\emph{etc}}\xspace}
\newcommand{\ie}{\hbox{\emph{i.e.}}\xspace}
\newcommand{\wrt}{\hbox{\emph{w.r.t.}}\xspace}

\newcommand{\tabincell}[2]{\begin{tabular}{@{}#1@{}}#2\end{tabular}}

\usepackage[most]{tcolorbox}

\title{Information Extraction in Low-Resource Scenarios: \\ Survey and Perspective}

\author{
\IEEEauthorblockN{Shumin Deng} 
\IEEEauthorblockA{\textit{National University of Singapore} \\
\textit{NUS-NCS Joint Lab}\\
Singapore \\
shumin@nus.edu.sg}
\and
\IEEEauthorblockN{Yubo Ma} 
\IEEEauthorblockA{\textit{S-Lab, Nanyang Technological University} \\
Singapore \\
yubo001@e.ntu.edu.sg}
\and
\IEEEauthorblockN{\quad\quad\quad Ningyu Zhang$^{\dagger}$\thanks{\quad$^{\dagger}$Corresponding Author.}\quad\quad\quad} 
\IEEEauthorblockA{\textit{Zhejiang University} \\
China \\
zhangningyu@zju.edu.cn}
\and
\IEEEauthorblockN{Yixin Cao} 
\IEEEauthorblockA{\textit{\quad\quad\quad\quad\quad\quad\quad Fudan University \quad\quad\quad\quad\quad\quad\quad} \\
China \\
yxcao@fudan.edu.cn}
\and
\IEEEauthorblockN{Bryan Hooi$^{\dagger}$} 
\IEEEauthorblockA{\textit{National University of Singapore} \\
\textit{NUS-NCS Joint Lab}\\
Singapore \\
dcsbhk@nus.edu.sg}
}

\begin{document}

\maketitle
\normalem

\begin{abstract}
Information Extraction (IE) seeks to derive structured information from unstructured texts, often encountering obstacles in low-resource scenarios due to data scarcity and unseen classes. 
This paper presents a review of neural approaches to low-resource IE from \emph{traditional} and \emph{LLM-based} perspectives, systematically organizing them into a fine-grained taxonomy. 
Our empirical studies compare LLM-based methods with prior leading models, revealing that: 
(1) well-tuned LMs perform relatively best;
(2) tuning open-resource LLMs and in-context learning with GPT family are generally effective;
(3) LLMs struggle to tackle complex tasks with intricate schema. 
Furthermore, we compare traditional methods and discuss LLM-based approaches, spotlighting promising applications and delineating future research directions. 
This survey aims to foster understanding of this field, inspire new ideas, and encourage widespread applications in both academia and industry. 
\end{abstract}

\section{Introduction}
\label{sec:intro}

{\ourslong} (\ours) \cite{J1996_IE,J2015_KE} offers essential support for a diverse range of domain-specific \cite{J2018_IE-Application,ICDE2023_OpenBG} and knowledge-intensive \cite{TNNLS2022_KE-Application,WWW2019-KGTA_IE4Finance} tasks, thus appealing to AI community. 
Common {\ours} tasks, including Named Entity Recognition (NER) \cite{TKDE2022_Survey_NER}, Relation Extraction (RE) \cite{J2022_Survey_RE}, and Event Extraction (EE) \cite{TNNLS2022_Survey_EE}, aim to derive structured data from unstructured texts. 
While deep learning has boosted {\ours} performance, it demands massive amounts of labeled data, which is difficult to acquire due to task and domain variability in practice. 
This issue motivates research in \textbf{low-resource \ours} (data-efficient \ours), which seeks to enhance learning efficiency under real world contexts with sparse data. 

Low-resource {\ours} has been widely investigated due to its potential for making models data-efficient and adaptable to various scenarios. 
Among numerous approaches proposed in recent years, Large Language Models (LLMs) \cite{arXiv2023_LLMSurvey,arXiv2023_LLMSurvey-Perspective} have demonstrated promising performance in low-resource scenarios. Hence, a timely review on low-resource {\ours} is beneficial in offering insights to both researchers and industry practitioners. 
Though several surveys on task-specific low-resource {\ours} (\eg, NER \cite{EMNLP2021_EmpStudy_FewShot-NER,J2023_Survey_FewShot_NER}, RE \cite{TACL2021_Survey_LowResRE,ACL2023_Survey_RE}, EE \cite{ACL2023_Survey_EE}) and low-resource NLP \cite{NAACL21_Survey_LowRes-NLP,TACL2023_Survey_EfficientNLP} have been published, a comprehensive survey for low-resource {\ours} encompassing both traditional models and modern LLMs is still absent. 
Therefore, in this paper, we provide a literature review on low-resource {\ours}, hoping to systematically analyse the methodologies, compare different technical solutions and inspire diverse new ideas. 

In this paper, 
we begin by introducing the basics of low-resource learning and {\ours} in $\S$\ref{sec:preliminary}.  
We then categorize low-resource {\ours} approaches\footnote{Due to page limits, complete papers are listed: \url{https://github.com/231sm/Low-resource-KEPapers}.} into \textbf{Traditional} ($\S$\ref{sec:slm_methods}) and \textbf{LLM-based Methods} ($\S$\ref{sec:llm_methods}).  
In view of \emph{which aspect is chiefly enhanced when training}, we further categorize \textbf{{\color{slm}{traditional low-resource {\ours} approaches}}} into three paradigms: 

\noindent
\emph{{\color{slm}{Exploiting Higher-Resource Data}}} ($\S$\ref{sec:slm_higher}): 
Increasing the size and diversity of the original sparse data with auxiliary resources which are generated endogenously or imported exogenously, aiming to obtain enriched samples and more precise semantic representations.

\noindent
\emph{{\color{slm}{Developing Stronger Data-Efficient Models}}} ($\S$\ref{sec:slm_stronger}): 
Developing more robust models to better cope with maldistribution of samples and new unseen classes, aiming to improve model learning abilities so as to reduce dependence on samples.

\noindent
\emph{{\color{slm}{Optimizing Data \& Models Together}}} ($\S$\ref{sec:slm_together}): 
Jointly utilizing representative samples and robust models, aiming at adapting to low-resource scenarios promptly, and searching more suitable strategies for learning with sparse data. 

We also divide \textbf{{\color{llm}{LLM-based low-resource {\ours} approaches}}} based on \emph{whether the LLM is tuned}, into two paradigms: 

\noindent
\emph{{\color{llm}{Direct Inference Without Tuning}}} ($\S$\ref{sec:llm_wo_tuning}): 
Prompting LLMs to generate answers with given instructions, where prompts can either be in textual or code format. 

\noindent
\emph{{\color{llm}{Model Specialization With Tuning}}} ($\S$\ref{sec:llm_w_tuning}):
Fine-tuning LLMs, either via prompt-tuning or standard fine-tuning. 
The fine-tuned LLMs are often tailored for specific {\ours} tasks.

We then conduct empirical studies comparing traditional and LLM-based low-resource {\ours} approaches in $\S$\ref{sec:experiments}. 
We explore the effectiveness of different techniques, especially those based on LLMs, to identify the most suitable approach for various low-resource {\ours} tasks.
The \textbf{key findings} are: 
(1) Well-tuned LMs continue to dominate in performance; 
(2) Tuning open-resource LLMs and using in-context learning (ICL) with GPTs presents promising results; 
(3) LLMs struggle to address the complex task with intricate schema. 
Besides, we summarize widely used benchmarks and promising applications in $\S$\ref{sec:datasets_app}. 
Finally, we compare existing low-resource {\ours} methods, discuss future directions in $\S$\ref{sec:compare_discuss}, and make a conclusion in $\S$\ref{sec:cons}.

\section{Preliminary on Low-resource IE}
\label{sec:preliminary}

\subsection{Information Extraction}
Generally, {\ours} \cite{EMNLP2020_OpenUE,EMNLP2022_DeepKE} can be regarded as structured prediction tasks \cite{ICML2005_StructPred}, where a classifier is trained to approximate a target function $F(x) \rightarrow y$, where $x \in \mathcal{X}$ denotes the input data and $y \in \mathcal{Y}$ denotes the output class sequence. This includes the subtasks of Named Entity Recognition (NER), Relation Extraction (RE)  and Event Extraction (EE).
For instance, given a sentence ``\emph{Jack is married to the Iraqi microbiologist known as Dr. Germ.}'': 

\noindent
\textbf{Named Entity Recognition} \cite{TKDE2022_Survey_NER} should identify the types of entities in given texts, \eg, ‘\emph{Jack}', ‘\emph{Dr. Germ}' $\Rightarrow$ \texttt{PERSON}; 

\noindent
\textbf{Relation Extraction} \cite{J2022_Survey_RE} should identify the relationship of the given entity pair $\langle$\emph{Jack}, \emph{Dr. Germ}$\rangle$ as \texttt{husband\_of};

\noindent
\textbf{Event Extraction} \cite{TNNLS2022_Survey_EE} should identify the event type as \texttt{Marry}, where the word ‘\emph{married}' triggers the event (subtask: ED, Event Detection); \emph{Jack} and \emph{Dr. Germ} are participants as \texttt{husband} and \texttt{wife} in the event respectively (subtask: EAE, Event Argument Extraction).

\subsection{Low-Resource Scenarios}

Most traditional {\ours} models \cite{TACL2016_NER_BiLSTM,EMNLP2015_RE_PCNN,ACL2015_EE_DMCNN} assume that sufficient training data are indispensable to achieve satisfactory performance. 
However, in real-world applications, task-specific labeled samples tend to be unevenly distributed and new unseen classes may emerge over time, which result in low-resource scenarios \cite{NAACL21_Survey_LowRes-NLP}. 
Considering \emph{maldistribution of samples} and \emph{new unseen classes}, we systematically categorize low-resource scenarios into three aspects. 

\noindent
\textbf{Long-tail Scenario} \cite{TPAMI2023_Long-Tail} means that only a minority of the classes are data-rich, while the majority of the classes have extremely few labeled sample data. 
Formally, given classes $\mathcal{Y} = \mathcal{Y}_h \cup \mathcal{Y}_t$ with  head classes ($\mathcal{Y}_h$) and long-tail classes ($\mathcal{Y}_t$), we denote the number of labeled samples for $\mathcal{Y}_h$ and $\mathcal{Y}_t$ as $|\mathcal{X}_h|$ and $|\mathcal{X}_t|$ respectively. $|\mathcal{X}_h|$ is much larger than $|\mathcal{X}_t|$, \ie, $|\mathcal{X}_h| \gg |\mathcal{X}_t|$, while $|\mathcal{Y}_h| \ll |\mathcal{Y}_t|$. 

\noindent
\textbf{Few-shot Scenario} \cite{J2020_Survey_FewLearning,PR2023_Survey_Few-ML} means that the classes for testing $\mathcal{Y}_{te}$ have only a small number of samples, where the small number can be \emph{fixed} following the N-Way-K-Shot setting \cite{NIPS2017_PN} (each of the N classes contains only K samples), or \emph{unfixed} and relatively small comparing with the total data size. 
Besides, $\mathcal{Y}_{te}$ is nonexistent (unseen) in the training dataset with classes $\mathcal{Y}_{tr}$, \ie, $\mathcal{Y}_{tr} \cap \mathcal{Y}_{te} = \emptyset$. 

\noindent
\textbf{Zero-shot Scenario} \cite{TIST2019_Survey_ZeroLearning} means that the testing classes $\mathcal{Y}_{te}$ to predict have no samples during the training phase. The model can classify these new unseen $\mathcal{Y}_{te}$ by utilizing prior knowledge or relationships between seen and unseen classes, represented as class attributes or embeddings.   
In \emph{standard zero-shot scenarios}, $\mathcal{Y}_{te}$ only contains unseen classes in the training set; 
in \emph{generalized zero-shot scenarios} \cite{TPAMI2023_Survey_GZSL}, $\mathcal{Y}_{te}$  contains both seen and unseen classes.

\tikzstyle{my-box}=[
    rectangle,
    draw=hidden-draw,
    rounded corners,
    text opacity=1,
    minimum height=1.5em,
    minimum width=5em,
    inner sep=2pt,
    align=center,
    fill opacity=.5,
    line width=0.8pt,
]
\tikzstyle{leaf}=[my-box, minimum height=1.5em,
    text=black, align=left,font=\normalsize, 
    inner xsep=2pt,
    inner ysep=4pt,
    line width=0.8pt,
]
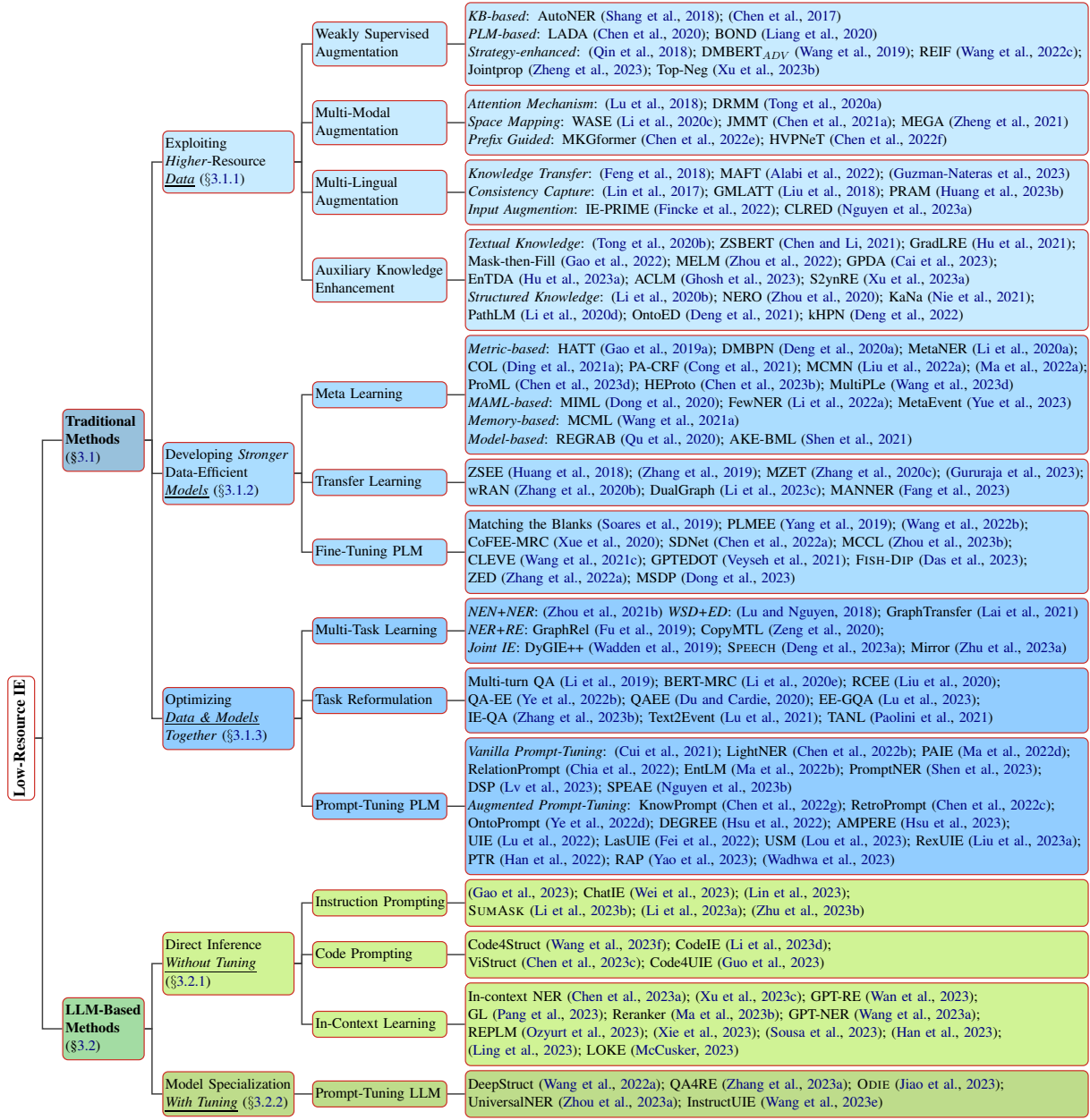
\begin{figure*}[!ht] 
    \centering
    \resizebox{\textwidth}{!}{
        \begin{forest}
            forked edges,
            for tree={
                grow=east,
                reversed=true,
                anchor=base west,
                parent anchor=east,
                child anchor=west,
                base=left,
                font=\large, 
                rectangle,
                draw=hidden-draw,
                rounded corners,
                align=left,
                minimum width=4em,
                edge+={darkgray, line width=1pt},
                s sep=3pt,
                inner xsep=2pt,
                inner ysep=3pt,
                line width=0.8pt,
                ver/.style={rotate=90, child anchor=north, parent anchor=south, anchor=center},
            },
            where level=1{text width=5.2em,font=\normalsize,}{},
            where level=2{text width=8.8em,font=\normalsize,}{},
            where level=3{text width=8.8em,font=\normalsize,}{},
            [
                \textbf{Low-Resource {\ours}}, ver
                [
                    \textbf{Traditional} \\ \textbf{Methods} \\ (\S\ref{sec:slm_methods})
                    ,fill=hidden-slm!70
                    [
                        Exploiting \\ \emph{Higher}-Resource \\ \underline{\emph{Data}} 
                        ($\S$\ref{sec:slm_higher})
                        , fill=hidden-slm1!70
                        [
                            Weakly Supervised \\ Augmentation
                            , fill=hidden-slm1!70
                            [
                                \emph{KB-based}: 
                                AutoNER \cite{EMNLP2018_DS-NER_AutoNER}; 
                                \cite{ACL2017_DS-EE} 
                                \\
                                \emph{PLM-based}: 
                                LADA \cite{EMNLP2020_DS-NER}; 
                                BOND \cite{KDD2020_DS-NER} 
                                \\
                                \emph{Strategy-enhanced}: 
                                \cite{ACL2018_DS-RE_Reinforcement}; %
                                DMBERT$_{ADV}$ \cite{NAACL2019_DS-EE}; 
                                REIF \cite{COLING2022_WS_RE}; 
                                Jointprop \cite{ACL2023_WS_RE}; 
                                Top-Neg \cite{ACL2023-Findings_DS-NER}
                                , leaf, text width=44em, fill=hidden-slm1!70
                            ]
                        ]
                        [
                            Multi-Modal \\ Augmentation
                            , fill=hidden-slm1!70
                            [
                                \emph{Attention Mechanism}: 
                                \cite{ACL2018_MM-NER};
                                DRMM \cite{AAAI2020_MM-EE} %
                                \\
                                \emph{Space Mapping}: 
                                WASE \cite{ACL2020_MM-EE}; 
                                JMMT \cite{EMNLP2021-Findings_MM-EE}; 
                                MEGA \cite{MM2021_MM-RE}
                                \\
                                \emph{Prefix Guided}: 
                                MKGformer \cite{SIGIR2022_MM-NER-RE};
                                HVPNeT \cite{NAACL2022-Findings_MM-NER-RE}
                                , leaf, text width=44em, fill=hidden-slm1!70
                            ]
                        ]
                        [
                            Multi-Lingual \\ Augmentation
                            , fill=hidden-slm1!70
                            [
                                \emph{Knowledge Transfer}: 
                                \cite{IJCAI2018_MultiL-NER}; 
                                MAFT \cite{COLING2022_MultiL-NER}; 
                                \cite{ACL2023_MultiL-EE} 
                                \\ 
                                \emph{Consistency Capture}:
                                \cite{ACL2017-MultiL-RE}; 
                                GMLATT \cite{AAAI2018_MultiL-EE}; %
                                PRAM \cite{ACL2023-Findings_MultiL-NER} 
                                \\
                                \emph{Input Augmention}:
                                IE-PRIME \cite{AAAI2022_MultiL-EE}; %
                                CLRED \cite{ACL2023-Findings_MultiL-EE}
                                , leaf, text width=44em, fill=hidden-slm1!70
                            ]
                        ]
                        [
                            Auxiliary Knowledge \\ Enhancement
                            , fill=hidden-slm1!70
                            [
                                \emph{Textual Knowledge}:
                                \cite{ACL2020_Know-EE};
                                ZSBERT \cite{NAACL2021_RE-ZSBERT}; 
                                GradLRE \cite{EMNLP2021_DA_RE}; 
                                Mask-then-Fill \cite{EMNLP2022_PLM-EE}; 
                                MELM \cite{ACL2022_PLM-NER}; 
                                GPDA \cite{ACL2023-Short_DA-NER}; \\ 
                                EnTDA \cite{ACL2023-Findings_DA-NER}; 
                                ACLM \cite{ACL2023_DataAug-NER_ACLM}; 
                                S2ynRE \cite{ACL2023_DA-RE_S2ynRE}
                                \\
                                \emph{Structured Knowledge}: 
                                \cite{COLING2020_Logic-RE}; 
                                NERO \cite{WWW2020_RE_NERO}; 
                                KaNa \cite{AAAI2021_Know-NER}; 
                                PathLM \cite{EMNLP2020_PathLM}; %
                                OntoED \cite{ACL2021_OntoED}; 
                                kHPN \cite{KBS2022_kHPN}
                                , leaf, text width=44em, fill=hidden-slm1!70
                            ]
                        ]
                    ]
                    [
                        Developing \emph{Stronger} \\ Data-Efficient \\ \underline{\emph{Models}} 
                        ($\S$\ref{sec:slm_stronger})
                        , fill=hidden-slm2!70
                        [
                            Meta Learning
                            , fill=hidden-slm2!70
                            [
                                \emph{Metric-based}: 
                                HATT \cite{AAAI2019_Meta-RE};
                                DMBPN \cite{WSDM2020_MetaL-EE_DMBPN}; 
                                MetaNER \cite{WWW2020_Meta-NER}; 
                                COL \cite{ICLR2021_Meta-RE_COL};
                                PA-CRF \cite{ACL2021-Finding_Meta-EE};
                                MCMN \cite{ACL2022_Meta-RE}; 
                                \cite{ACL2022-Findings_Meta-NER_MetricBased}; \\
                                ProML \cite{ACL2023-Findings_Meta-NER};
                                HEProto \cite{CIKM2023_Meta-NER};
                                MultiPLe \cite{CIKM2023_Meta-EE}
                                \\
                                \emph{MAML-based}: 
                                MIML \cite{COLING2020_Meta-RE};
                                FewNER \cite{TKDE2022_Meta-NER};
                                MetaEvent \cite{ACL2023_Meta-EE}
                                \\
                                \emph{Memory-based}: 
                                MCML \cite{arXiv2023_Meta-NER_MCML}
                                \\
                                \emph{Model-based}: 
                                REGRAB \cite{ICML2020_Meta-RE}; 
                                AKE-BML \cite{ACL2021-Findings_Meta-EE_AKE-BML}
                                , leaf, text width=44em, fill=hidden-slm2!70
                            ]
                        ]
                        [
                            Transfer Learning
                            , fill=hidden-slm2!70
                            [
                                ZSEE \cite{ACL2018_ZSEE}; 
                                \cite{NAACL2019_Tranfer-RE}; 
                                MZET \cite{COLING2020_Tranfer-NER}; 
                                \cite{ACL2023_Tranfer-RE}; 
                                wRAN \cite{WWW2020_Tranfer-RE_wRAN};  
                                DualGraph \cite{TNNLS2023_Tranfer-RE_DualGraph}; 
                                MANNER \cite{ACL2023_Tranfer-NER_MANNER}
                                , leaf, text width=44em, fill=hidden-slm2!70
                            ]
                        ]
                        [
                            Fine-Tuning PLM
                            , fill=hidden-slm2!70
                            [
                                Matching the Blanks \cite{ACL2019_RE-MTB}; 
                                PLMEE \cite{ACL2019_Pretraning-EE-PLMEE}; 
                                \cite{ACL2022-Findings_Pretraning-EE}; 
                                CoFEE-MRC \cite{EMNLP2020_Pretraning-NER}; 
                                SDNet \cite{ACL2022_Pretraning-NER}; 
                                MCCL \cite{ACL2023_Pretraning-RE_MCCL}; \\ 
                                CLEVE \cite{ACL2021_Pretraning-EE_CLEVE}; 
                                GPTEDOT \cite{ACL2021_GPT-2_EE}; 
                                ClarET \cite{ACL2022_Pretraning-EE_ClarET}; 
                                \textsc{Fish-Dip} \cite{EMNLP2023_Pretraning-TE_FISH-DIP}; 
                                ZED \cite{EMNLP2022-Findings_Pretraining-EE}; %
                                MSDP \cite{CIKM2023_Pretraning-NER} %
                                , leaf, text width=44em, fill=hidden-slm2!70
                            ]
                        ]
                    ]
                    [
                        Optimizing \\ \underline{\emph{Data \& Models}} \\ \emph{Together} 
                        ($\S$\ref{sec:slm_together})
                        , fill=hidden-slm3!70
                        [
                            Multi-Task Learning,
                             fill=hidden-slm3!70
                            [
                                \emph{NEN+NER}: \cite{ACL2021_MTL-NER} 
                                \emph{WSD+ED}: \cite{EMNLP2018_MTL-EE}; GraphTransfer \cite{SIGIR2021_MTL-EE} 
                                \\
                                \emph{NER+RE}: GraphRel \cite{ACL2019_MTL-RE};
                                CopyMTL \cite{AAAI2020_MTL-RE}; 
                                \\ 
                                \emph{Joint {\ours}}: DyGIE++ \cite{EMNLP2019_MTL-KE}; 
                                \textsc{Speech} \cite{ACL2023_SPEECH}; 
                                Mirror \cite{EMNLP2023_MTL-IE}
                                , leaf, text width=44em, fill=hidden-slm3!70
                            ]
                        ]
                        [
                            Task Reformulation
                            , fill=hidden-slm3!70
                            [
                                Multi-turn QA \cite{ACL2019_QA-NER-RE};  
                                BERT-MRC \cite{ACL2020_QA-NER}; 
                                RCEE \cite{EMNLP2020_RCEE}; 
                                QAEE \cite{EMNLP2020_QAEE}; 
                                EE-GQA \cite{ACL2023-Short_EE-GQA}; 
                                IE-QA \cite{TASLP2023_IE-QA}; 
                                \\
                                Text2Event \cite{ACL2021_Text2Event}; 
                                TANL \cite{ICLR2021_TANL}; 
                                Set \cite{EMNLP2023_GenIE_Set} 
                                , leaf, text width=44em, fill=hidden-slm3!70
                            ]
                        ]
                        [
                            Prompt-Tuning PLM
                            , fill=hidden-slm3!70
                            [
                                \emph{Vanilla Prompt-Tuning}:
                                \cite{ACL2021_Prompt-NER}; 
                                LightNER \cite{COLING2022_Prompt-NER}; 
                                PAIE \cite{ACL2022_Prompt-EAE}; 
                                RelationPrompt \cite{ACL2022-Finding_RelationPrompt}; 
                                EntLM \cite{NAACL2022_Prompt-NER}; 
                                \cite{ACL2022_Prompt-NER}; 
                                \\
                                PromptNER \cite{ACL2023_Prompt-NER_PromptNER}; 
                                DSP \cite{ACL2023-Findings_Prompt-RE}; 
                                SPEAE \cite{ACL2023-Findings_Prompt-EE} 
                                \\
                                \emph{Augmented Prompt-Tuning}:
                                KnowPrompt \cite{WWW2022_Prompt-RE}; 
                                RetroPrompt \cite{NeurIPS2022_Retrieval-RetroPrompt}; 
                                OntoPrompt \cite{WWW2022_Prompt-EE}; 
                                DEGREE \cite{NAACL2022_Prompt-EE}; \\
                                AMPERE \cite{ACL2023_Prompt-EE_AMPERE}; 
                                UIE \cite{ACL2022_UIE}; 
                                LasUIE \cite{NeurIPS2022_LasUIE}; 
                                USM \cite{AAAI2023_USM}; 
                                RexUIE \cite{EMNLP2023-Findings_Prompt-IE_RexUIE};  
                                PTR \cite{J2022_Prompt-RE};
                                RAP \cite{SIGIR2023_Prompt-KE_RAP};
                                \cite{ACL2023_Survey_RE}
                                , leaf, text width=44em, fill=hidden-slm3!70
                            ]
                        ]
                    ]
                ]
                [
                    \textbf{LLM-Based} \\ \textbf{Methods} \\ (\S\ref{sec:llm_methods})
                    , fill=hidden-llm!70
                    [ 
                        Direct Inference \\ \underline{\emph{Without Tuning}} \\ 
                        ($\S$\ref{sec:llm_wo_tuning})
                        , fill=hidden-llm1!70
                        [
                            Instruction Prompting
                            , fill=hidden-llm1!70
                            [
                                \cite{arXiv2023_LLM4EE}; 
                                ChatIE \cite{arXiv2023_ChatIE}; 
                                \cite{EACL2023-Findings_InstructionPrompting-EE}; 
                                \textsc{SumAsk} \cite{EMNLP2023-Findings_InstructionPrompting_RE}; 
                                PromptNER \cite{arXiv2023_InstructionPrompting_PromptNER}; 
                                \cite{arXiv2023_LLM4IE};  
                                AutoKG \cite{arXiv2023_LLM4KGCR}
                                , leaf, text width=44em, fill=hidden-llm1!70
                            ]
                        ]
                        [
                            Code Prompting
                            , fill=hidden-llm1!70
                            [
                                Code4Struct \cite{ACL2023_Code4Struct}; 
                                CodeIE \cite{ACL2023_CodeIE}; 
                                CodeKGC \cite{arXiv2023_CodeKGC}; 
                                ViStruct \cite{EMNLP2023_ViStruct}; 
                                Code4UIE \cite{arXiv2023_RA-CodeIE}; 
                                GoLLIE \cite{arXiv2023_GoLLIE}
                                , leaf, text width=44em, fill=hidden-llm1!70
                            ]
                        ]
                        [
                            In-Context Learning
                            , fill=hidden-llm1!70
                            [
                                In-context NER \cite{ACL2023_InContextNER}; 
                                \cite{ACL2023-SustailNLP_Survey_RE}; 
                                GPT-RE \cite{EMNLP2023_GPT-RE}; 
                                GL \cite{EMNLP2023_ICL_IE_Agent}; 
                                Reranker \cite{EMNLP2023-Findings_LLM4LowResKE}; 
                                GPT-NER \cite{arXiv2023_GPT-NER}; \\
                                REPLM \cite{arXiv2023_ICL_RE};
                                \cite{arXiv2023_ICL_NER}; 
                                \cite{arXiv2023_ICL_IE}; 
                                \cite{arXiv2023_LLM4IE_2}; 
                                \cite{arXiv2023_LLM4OpenIE}; 
                                LOKE \cite{arXiv2023_LLM4OpenKGC};
                                \cite{EMNLP2023-Findings_ICL-MNER}; 
                                PGIM \cite{EMNLP2023-Findings_ICL-MNER_PGIM}
                                , leaf, text width=44em, fill=hidden-llm1!70
                            ]
                        ]   
                    ]
                    [
                        Model Specialization \\ \underline{\emph{With Tuning}} 
                        ($\S$\ref{sec:llm_w_tuning})
                        , fill=hidden-llm2!70
                        [
                            Prompt-Tuning LLM
                            , fill=hidden-llm2!70
                            [
                                DeepStruct \cite{ACL2022-Findings_DeepStruct}; 
                                QA4RE \cite{ACL2023-Findings_PromptTuning_RE}; 
                                \textsc{Odie} \cite{EMNLP2023_PromptTuning_IE}; 
                                UniversalNER \cite{ICLR2024_PromptTuning_UniversalNER};
                                InstructUIE \cite{arXiv2023_PromptTuning_InstructUIE}
                                , leaf, text width=44em, fill=hidden-llm2!70
                            ]
                        ]
                    ]
                ]
            ]
        \end{forest}
    }
    \caption{
    Taxonomy of low-resource {\ours} methods. (We only list part of representative approaches, referring to the complete paper list$^{\text{1}}$ for more details.)
    \label{fig:taxonomy}
    }
    
\end{figure*}

\section{Taxonomy of Technical Solutions}
\label{sec:method}

In general, for this survey, we select influential low-resource {\ours} papers mostly published within the last five years. The majority of these papers are sourced from prestigious conferences and journals, especially in NLP and machine learning domains, \eg, ACL, EMNLP, NAACL, NeurIPS, ICLR, TACL, TKDE and so on. 
We generally categorize existing approaches into {\color{slm}{traditional}} and {\color{llm}{LLM-based}} methods based on the \emph{model parameter size} with a threshold: 6B. 
For \textbf{Traditional Methods}, the backbone model parameter size is less than 6B, \eg, BERT \cite{NAACL2019_BERT}, BART \cite{ACL2020_BART}, T5-large \cite{JMLR2020_T5}, GPT-2 \cite{OpenAI2019_GPT2}, GPT-J\footnote{\url{https://huggingface.co/EleutherAI/gpt-j-6b}.}, \etc.  
For \textbf{LLM-based Methods}, the backbone model parameter size is larger than 6B, \eg, FlanT5-11B \cite{arXiv2022_FlanT5}, LLaMA \cite{MetaAI2023_LLaMA}, GLM \cite{ACL2022_GLM}, GPT-3 \cite{NeurIPS2020_GPT3}, ChatGPT\footnote{\url{https://openai.com/blog/chatgpt}.}, GPT-4 \cite{OpenAI2023_GPT4}, \etc. 
We refine representative methods in the two categories, and establish a fine-grained taxonomy as presented in Figure~\ref{fig:taxonomy}. 

\subsection{Traditional Methods}
\label{sec:slm_methods}

In view of \emph{which aspect is chiefly enhanced when addressing low-resource {\ours} problems}, we categorize some representative {\color{slm}{traditional methods}} into three general paradigms, as shown in Figure~\ref{fig:taxonomy}. 
In summary, traditional low-resource {\ours} models tend to 
(1) exploit \emph{higher}-resource data; 
(2) develop \emph{stronger} data-efficient models; and
(3) optimize data \& models \emph{together}. 

\subsubsection{Exploiting Higher-Resource Data}
\label{sec:slm_higher}
This paradigm mainly refers to data augmentation \cite{TACL2023_DAinNLP} or knowledge enhancement on the original small dataset, with endogenous or exogenous auxiliary resources. 
\emph{The goal is to create more representative samples and improve semantic representations with higher-resource data.} 

\noindent
\textbf{Weakly Supervised Augmentation} involves synthesizing more training data through weak/distant supervision. 
This kind of methods usually utilize a knowledge base (KB) and some heuristic rules to automatically relabel training instances in corpus, potentially resulting in a noisy synthesized dataset. 

\cite{ACL2009_DS-RE} proposed \emph{distant supervision} for RE, utilizing a large semantic KB --- Freebase to label relations in an unlabeled corpus. 
Similarly, \cite{EMNLP2018_DS-NER_AutoNER}  used a dictionary for distantly supervised NER. \cite{ACL2017_DS-EE} leveraged Freebase and FrameNet (an event KB) to automatically label training data for EE. 
Recent studies have proposed to improve quality of weakly supervised data with pretrained LMs (PLMs) \cite{EMNLP2020_DS-NER,KDD2020_DS-NER} and optimization strategies \cite{ACL2018_DS-RE_Reinforcement,NAACL2019_DS-EE,COLING2022_WS_RE,ACL2023_WS_RE,ACL2023-Findings_DS-NER}. 
To mitigate selection bias, \cite{EMNLP2021_DA_RE} have proposed a gradient imitation reinforcement learning framework for low-resource RE. 

\noindent
\textbf{Multi-Modal Augmentation} supplements single modal with multi-modal samples to enhance semantics and facilitate disambiguation. 
Intuitively, the main challenge of such methods lies in effectively fusing data from different modalities. 

%
To fuse multi-modal data in low-resource {\ours}, \cite{ACL2018_MM-NER,AAAI2020_MM-EE} utilized attention mechanisms, \cite{ACL2020_MM-EE,EMNLP2021-Findings_MM-EE,MM2021_MM-RE} imposed multi-modal embedding space mapping, and \cite{SIGIR2022_MM-NER-RE,NAACL2022-Findings_MM-NER-RE} leveraged prefix-guided multi-modal fusion. 

\noindent
\textbf{Multi-Lingual Augmentation} involves incorporating multi-lingual samples to achieve diverse and robust sample representation. 
Intuitively, the main challenge of such methods is to obtain language representations cross linguistics. 

%
To tackle the issue, transferring cross-lingual knowledge \cite{IJCAI2018_MultiL-NER,COLING2022_MultiL-NER,ACL2023_MultiL-EE}
and capturing the consistency cross languages \cite{ACL2017-MultiL-RE,AAAI2018_MultiL-EE,J2022_MultiL-RE,ACL2023-Findings_MultiL-NER} demonstrated effectiveness. 
Other promising models are to import additional context \cite{ACL2023-Findings_MultiL-EE} and task-specific knowledge \cite{AAAI2022_MultiL-EE} for cross-lingual {\ours}. 

\noindent
\textbf{Auxiliary Knowledge Enhancement} employs external knowledge as remedies, intended to learn semantic representation of samples more precisely. 
Different from weakly supervised augmentation using KB, this paradigm adopts more diverse knowledge formats and  knowledge enhancement methods. 
Given diverse formats of auxiliary knowledge, we divide them into two categories. 

\noindent
\emph{(i) Textual Knowledge} 

Textual knowledge can be \emph{class-related knowledge} \cite{ACL2019_Know-EE,EMNLP2021_RE-MapRE}, like class descriptions \cite{EMNLP2018_Know-NER_ZOE,NAACL2019_Know-NER_DZET,NAACL2021_RE-ZSBERT} and class-specific texts \cite{ACL2020_Know-EE}. 
It can also be \emph{synthesized data} \cite{ACL2022_PLM-NER,EMNLP2022_PLM-EE,IJCAI2022_DA-NER_PromptDA,ACL2023_DataAug-NER_ACLM,ACL2023-Findings_DA-NER,ACL2023-Short_DA-NER,ACL2023-Findings_DA-RE_GDA,ACL2023-Findings_DA-RE_GPT2Training,ACL2023_RE-REMatching,ACL2023_DA-RE_S2ynRE,arXiv2023_SegMix_IE} via data augmentation. 

\noindent
\emph{(ii) Structured Knowledge} 
Structured knowledge for low-resource {\ours} can take the form of KG triples \cite{ACL2016_Know-EE}, task-specific ontology \cite{SIGIR2021_Know-NER}, and rules \cite{WWW2020_RE_NERO}. 
For example, \cite{COLING2020_Logic-RE} studied zero-shot RE by leveraging KG embeddings and logic rules to connect seen and unseen relations; 
\cite{AAAI2021_Know-NER} proposed a KB-aware NER framework to utilize class-heterogeneous knowledge in KBs; 
\cite{EMNLP2020_PathLM,ACL2021_OntoED,KBS2022_kHPN} resolved low-resource EE problems by utilizing association knowledge among classes. 


\subsubsection{Developing Stronger Data-Efficient Models}
\label{sec:slm_stronger}
The paradigm focuses on developing models to handle sample maldistribution and unseen classes more effectively. 
\emph{The stronger models aim to improve learning abilities, so as to maximize the utilization of small data and minimize the dependence of samples.} 

\noindent
\textbf{Meta Learning} \cite{TPAMI2022_MetaLearning} promptly assimilates emerging knowledge and deduces new classes by learning from few instances, with the ability of ``learning to learn'', which is naturally suitable for few-shot {\ours} tasks.  

\cite{AAAI2019_Meta-RE,WSDM2020_MetaL-EE_DMBPN,WWW2020_Meta-NER,ICLR2021_Meta-RE_COL,ACL2021-Finding_Meta-EE,ACL2022_Meta-RE,ACL2022-Findings_Meta-NER_MetricBased,ACL2023-Findings_Meta-NER,CIKM2023_Meta-NER,CIKM2023_Meta-EE} utilized \emph{metric-based} methods, mostly equipped with prototypical networks \cite{NIPS2017_PN}. 
\cite{ACL2019_Meta-RE_MLRC,COLING2020_Meta-RE,AAAI2020-SA_Meta-IE,TKDE2022_Meta-NER,ACL2022-Findings_Meta-NER,ACL2023_Meta-EE} leveraged \emph{model-agnostic} methods based on MAML \cite{ICML2017_MAML}. 
\cite{arXiv2023_Meta-NER_MCML} proposed a \emph{memory-based} method \cite{ICML2016_MANN}. 
\cite{ICML2020_Meta-RE,ACL2021-Findings_Meta-EE_AKE-BML} used \emph{model-based} methods with Bayesian meta learning \cite{NeurIPS2020_BayesianMetaLearning}.

\noindent
\textbf{Transfer Learning} \cite{TKDE2010_Tranfer-Survey} reduces the dependence on labeled target data by transferring learned class-invariant features, especially from high-resource to low-resource classes. 
 
\cite{ACL2018_ZSEE} leveraged class structures to transfer knowledge from existing to unseen classes. 
\cite{NAACL2019_Tranfer-RE} proposed a weighted adversarial network to adapt features learned from high-resource to low-resource classes. 
\cite{NAACL2019_Tranfer-RE,TNNLS2023_Tranfer-RE_DualGraph} utilized graph neural networks to facilitate knowledge transfer. 
\cite{ACL2023_Tranfer-NER_MANNER,COLING2020_Tranfer-NER} used a memory module for information retrieval or similarity comparison from the source to target domain. 
\cite{ACL2023_Tranfer-RE} incorporated linguistic representations for cross-domain few-shot transfer. 

\noindent
\textbf{Fine-Tuning PLM} leverages PLMs \cite{J2021_PLMSurvey} to utilize contextual representations and pre-trained parameters for fine-tuning. It adapts PLMs' powerful language understanding capabilities to specific low-resource {\ours} tasks. 

Fine-tuning PLMs for low-resource {\ours} aims at learning task-specific 
entity representations \cite{EMNLP2020_Pretraning-NER,ACL2022_Pretraning-NER,CIKM2023_Pretraning-NER}; 
relation representations \cite{ACL2019_RE-MTB,ACL2023_Pretraning-RE_MCCL,EMNLP2023_Pretraning-TE_FISH-DIP}; and 
event representations \cite{ACL2019_Pretraning-EE-PLMEE,ACL2021_GPT-2_EE,ACL2021_Pretraning-EE_CLEVE,ACL2022-Findings_Pretraning-EE,EMNLP2022-Findings_Pretraining-EE,ACL2022_Pretraning-EE_ClarET} with data-efficient learning. 


\subsubsection{Optimizing Data \& Models Together}
\label{sec:slm_together}
This paradigm refers to jointly optimizing representative samples and data-efficient models, enabling swift adaptation to low-resource scenarios. 
\emph{The aim is to identify more suitable strategies for learning with sparse data.}

\noindent
\textbf{Multi-Task Learning} signifies learning multiple related tasks simultaneously by exploiting both task-generic commonality and task-specific diversity \cite{ACL2022-Findings_DeepStruct}, contributing to improved performance for task-specific models, which naturally enables boost the target low-resource {\ours} task.

\noindent
\emph{(i) {\ours} \& \ours-Related Tasks}

\cite{ACL2021_MTL-NER,AAAI2019_MTL-NER,AAAI2021_MTL-NER} proposed to jointly model NER and Named Entity Normalization (NEN), as these two tasks can benefit each other with enhanced entity mention features. 
\cite{EMNLP2018_MTL-EE,SIGIR2021_MTL-EE} proposed to transfer the knowledge learned on Word Sense Disambiguation (WSD) to ED (a subtask of EE), considering ED and WSD are two similar tasks in which they both involve identifying the classes (\ie, event types or word senses) of some words in a given sentence. 

\noindent
\emph{(ii) Joint {\ours} \& Other Structured Prediction Tasks}

Different {\ours} and structured prediction \cite{ICML2005_StructPred} tasks can also benefit from each other considering the similar task structures and progressive task process. For example, 
\cite{ACL2019_MTL-RE,AAAI2020_MTL-RE} tackled the joint NER and RE task, respectively leveraging the relational graph and a copy mechanism;  
\cite{EMNLP2019_MTL-KE} incorporated global context in a general multi-task {\ours} framework, \wrt the NER, RE and EE;  
\cite{ACL2023_SPEECH,EMNLP2023_MTL-IE} jointly addressed {\ours} and other structured prediction tasks, such as event-relation extraction, multi-span extraction, and n-ary tuple extraction. 

\noindent
\textbf{Task Reformulation} refers to formulating {\ours} tasks into other formats which imports task-related knowledge, to capitalize on model architecture and data advantages. 
For example, {\ours} can be reformulated as \emph{machine reading comprehension (MRC)} or \emph{text-to-structure generation} tasks. 
MRC-based {\ours} identifies answer spans in context of questions, providing crucial knowledge for target tasks. 
Generative {\ours} \cite{NAACL2022_GenIE,EMNLP2022_GenerativeKGC,arXiv2023_GenerativeIE} employs generative LMs (GenLMs) for {\ours}, which can reduce error propagation while increasing adaptability for {\ours}. 

\noindent
\emph{(i) Reformulating {\ours} as QA/MRC.} 
\cite{EMNLP2020_RCEE} tackled low-resource EE by transferring event schema into natural questions.  
\cite{EMNLP2020_QAEE,AAAI2022-SA_QA-EE,ACL2023-Short_EE-GQA} studied how question generation strategies affect QA-based EE. 
\cite{ACL2019_QA-NER-RE,ACL2020_QA-NER,EACL2023_RE-QA} explored QA-based {\ours} to efficiently encode of crucial information of classes.

\noindent
\emph{(ii) Reformulating {\ours} as Text-to-Structure Generation.} 
\cite{ACL2021_Text2Event} proposed a sequence-to-structure generation paradigm for EE. 
\cite{ICLR2021_TANL} framed {\ours} as a translation task, efficiently extracting task-relevant information. 
\cite{EMNLP2023_GenIE_Set} employed sequence-to-sequence {\ours} with set learning to reduce structure order bias. 

\noindent
\textbf{Prompt-Tuning PLM} \cite{J2023_Prompt-Survey} involves inserting text pieces, \ie, templates, into the input to convert a classification task into a masked language modeling problem. 
This enables {\ours} approaches to benefit from LMs' pretrained knowledge, improving sample efficiency. 

\noindent
\emph{(i) Vanilla Prompt Tuning} 

Vanilla prompt tuning methods leverage the basic prompt learning framework, excelling in low-resource scenarios. 
\cite{ACL2021_Prompt-NER,NAACL2022_Prompt-NER,ACL2023_Prompt-NER_PromptNER,ACL2022_Prompt-NER,COLING2022_Prompt-NER,COLING2022_Prompt-NER_COPNER} respectively applied a template-based, template-free, dynamic-template-filling, demonstration-based, lightweight and contrastive prompt-based method for few-shot NER. 
\cite{ACL2022-Finding_RelationPrompt,ACL2023-Findings_Prompt-RE} leveraged the structured template and discriminative soft prompts for zero-shot RE. 
\cite{ACL2022_Prompt-EE,ACL2022_Prompt-EAE,ACL2023-Findings_Prompt-EE,ACL2023-Short_Prompt-EE_TypePrompt} respectively utilized template-based, extractive, contextualized, and type-specific prompts for low-resource EE. 

\noindent
\emph{(ii) Augmented Prompt Tuning} 

Augmented prompt tuning methods enhance vanilla prompt learning by incorporating diverse knowledge, facilitating low-resource {\ours}. 
\cite{WWW2022_Prompt-RE} incorporated knowledge among labels to prompt-tuning. 
\cite{NAACL2022_Prompt-EE,SIGIR2023_Prompt-KE_RAP} enhanced prompts with label semantics and class description. 
\cite{WWW2022_Prompt-EE,J2022_Prompt-RE,ACL2023_Prompt-EE_AMPERE,ACL2023_Survey_RE} demonstrated the effectiveness of enhancing prompt-tuning with ontology, rules, semantic structures and reasoning rationales. 
\cite{ACL2022_UIE,AAAI2023_USM} introduced a unified text generation (UIE) and semantic matching (USM) framework for different {\ours} tasks. 
\cite{NeurIPS2022_LasUIE} proposed a structure-aware GenLM to harness syntactic knowledge for UIE. 
\cite{EMNLP2023-Findings_Prompt-IE_RexUIE} introduced UIE for any kind of schemas. 
\cite{SIGIR2022_Retrieval-RetrieRE,EMNLP2022_Retrieval-RGQA,NeurIPS2022_Retrieval-RetroPrompt,TKDE2023_Retrieval-EE} utilized
 retrieval-augmented prompts to import task-specific knowledge. 

\subsection{LLM-Based Methods}
\label{sec:llm_methods}

Comparing with traditional PLMs, LLMs possess more powerful pretrained abilities, allowing for more complex prompt learning. 
In view of \emph{whether the LLM is tuned (i.e., LLM initial parameters are modified)}, we categorize some representative {\color{llm}{LLM-based methods}} into two general paradigms, as shown in Figure~\ref{fig:taxonomy}.

\subsubsection{Direct Inference without Tuning}
\label{sec:llm_wo_tuning}

This kind of methods typically involve ways to leverage LLMs without extensive additional training. 
They can boost low-resource {\ours} by leveraging the inherent capabilities of LLMs to understand and process contexts, further obtaining valuable insights from scarce data, thereby reducing the requirement of fine-tuning. 

\noindent
\textbf{Instruction Prompting} involves giving the LLM explicit instructions (without demonstrations) to perform a specific task. For low-resource {\ours}, instruction prompting can be effective as it allows the model to execute tasks using its pre-existing knowledge and language understanding. 

As instruction prompting does not require demonstrations, it is naturally suitable for zero-shot \cite{ACL2023-BioNLP_InstructionPrompting_RE,arXiv2023_ChatIE,arXiv2023_LLM4IE,arXiv2023_LLM4KGCR} and cross-domain \cite{arXiv2023_InstructionPrompting_PromptNER} {\ours} tasks. \cite{arXiv2023_LLM4EE} investigated zero-shot ED task and found that ChatGPT is competitive in simple scenarios, but struggles in more complex and long-tail scenarios. \cite{EACL2023-Findings_InstructionPrompting-EE} used global constraints with prompting for zero-shot EE, demonstrating adaptability to any other datasets. 
\cite{EMNLP2023_Survey_ZS-NER,EMNLP2023-Findings_InstructionPrompting_RE} observed that ChatGPT is advantageous in zero-shot NER and RE tasks with instruction prompting. 

\noindent
\textbf{Code Prompting} involves presenting the LLM with snippets of code (or code-like instructions \cite{arXiv2023_GoLLIE}) to guide its responses. This kind of methods can be particularly useful in low-resource {\ours} tasks that involve structured output, as the code implies the schema of the specific task. 

\cite{ACL2023_Code4Struct,ACL2023_CodeIE,arXiv2023_CodeKGC} tackled low-resource {\ours} tasks with code prompting on Code-LLMs, such as \textsc{CodeX} \cite{arXiv2021_CodeX}, demonstrating the effectiveness of code-style prompts. 
\cite{arXiv2023_RA-CodeIE} proposed a universal retrieval-augmented code generation framework. 
Moreover, code prompting can also be applied to multimodal {\ours} tasks \cite{EMNLP2023_ViStruct}. 

\noindent
\textbf{In-Context Learning (ICL)} utilizes the ability of LLMs to learn from the context provided in prompts. The model uses few relevant examples (demonstrations) to ``understand'' the specific {\ours} task and then applies this understanding to new data, particularly useful in low-resource scenarios. 

Recent research leveraged ICL for low-resource NER \cite{ACL2023_InContextNER,EMNLP2023-Findings_ICL-MNER,EMNLP2023-Findings_ICL-MNER_PGIM,arXiv2023_GPT-NER,arXiv2023_ICL_NER}, RE \cite{ACL2023-SustailNLP_Survey_RE,EMNLP2023_GPT-RE,EMNLP2023-Findings_ICL-RE,arXiv2023_ICL_RE}, joint {\ours} \cite{EMNLP2023-Findings_LLM4LowResKE,EMNLP2023_ICL_IE_Agent,arXiv2023_LLM4IE_2,arXiv2023_ICL_IE} and OpenIE \cite{arXiv2023_LLM4OpenIE,arXiv2023_LLM4OpenKGC}. 
\emph{The key challenges of ICL for {\ours} are: 
(1) input prompting fails to thoroughly express intricate {\ours} tasks, and 
(2) aligning input and labels is not effective enough. 
To tackle these issues, synthesizing data with LLMs \cite{EMNLP2023_SynthIE,EMNLP2023-Findings_LLMAnnotator} and inputting more task-specific prompts are promising.} 

\subsubsection{Model Specialization with Tuning}
\label{sec:llm_w_tuning}

This kind of methods enhance low-resource {\ours} by tailoring the model capabilities to specific tasks, and can be divided into prompt-tuning and fine-tuning. 
Generally, prompt-tuning is particularly valuable for its efficiency and minimal data requirements. 
Fine-tuning, while more resource-intensive, offers deeper customization and potentially better performance.

\noindent
\textbf{Prompt-Tuning LLM} involves keeping the LLM weights fixed and only tuning a small set of parameters associated with the prompts. 
In low-resource {\ours} tasks, prompt-tuning allows for adapting the model to specific tasks or domains with minimal data, where prompts act as a guide. 

\cite{arXiv2023_PromptTuning_InstructUIE} introduced a benchmark of diverse {\ours} tasks with expert-written instructions, and proposed a unified {\ours} framework, InstructUIE, with instruction tuning on FlanT5-11B \cite{arXiv2022_FlanT5}. 
\cite{EMNLP2023_PromptTuning_IE} proposed InstructIE dataset and finetune LLaMA-7B \cite{MetaAI2023_LLaMA} with instruction-following capability on {\ours}. 
\cite{ACL2023-Findings_PromptTuning_RE} hypothesized that instruction-tuning has been unable to elicit strong RE capabilities in LLMs by aligning RE with QA. 
\cite{ACL2022-Findings_DeepStruct} pretrained GLM-10B \cite{ACL2022_GLM} on a collection of task-agnostic corpora to generate structures from text. 
\cite{ICLR2024_PromptTuning_UniversalNER} explored targeted distillation with mission-focused instruction tuning to train student models on NER task.

\noindent
\textbf{Fine-Tuning LLM} involves adjusting the weights of the LLM on a smaller and task-specific dataset. This approach is more data-intensive than prompt-tuning but expectantly leads to performance gains. 

Actually, this kind of methods are currently underdeveloped for low-resource {\ours} tasks due to the limit of sufficient computing resources. 

\begin{table*}[!t] 
\centering
\small


\resizebox{\linewidth}{!}{

\begin{tabular}{c | l | c c c | c | c | c | c | c | c | c}

\toprule

\multirow{4}*{\textbf{Task}} & 
\multirow{4}*{\textbf{Dataset}} & 
\multicolumn{3}{c|}{\textbf{}} & 
\multicolumn{3}{c|}{\textbf{Prompt-Tuning/Fine-Tuning LLM}} &
\multicolumn{4}{c}{\textbf{In-Context Learning with LLM}}
\\ 
& & 
\multicolumn{3}{c|}{\textbf{Previous SOTA}} & 
\textbf{InstructUIE} & 
\textbf{KnowLM} & 
\textbf{FT ChatGPT} & 
\textbf{Vicuna} & 
\textbf{Vicuna 1.5} & 
\textbf{ChatGPT} & 
\textbf{GPT-4} 
\\ 
& & full & 0-shot & 5-shot 
& 0-shot test & 0-shot test & 0-shot test
& 5-shot ICL & 5-shot ICL & 5-shot ICL & 5-shot ICL
\\
& & & all-way | 10-way & all-way | 10-way 
& all-way & all-way & all-way & all-way & all-way & all-way & all-way \\

\midrule

\multirow{3}*{NER} 
& CoNLL03 		& 94.60 & $^\triangleright$74.99 | \quad-\quad~~ & 83.25 | \quad-\quad\quad 
& \underline{\textbf{92.94}} & 92.65 & 75.26 
& 34.00 & 44.40 & 77.80 & \uuline{\textbf{84.89}} \\ 

& OntoNotes5.0  & 92.30 & - & 59.70 | \quad-\quad\quad 
& \underline{\textbf{90.19}} & 80.40 & 55.34 
& 22.38 & 26.38 & 59.39 & \uuline{\textbf{68.71}} \\ 

& FewNERD 		& 70.90 & - &  59.41 | 79.00 
& 35.41 & \underline{\textbf{71.47}} & 57.22 
& 23.30 & 27.43 & 56.03 & \uuline{\textbf{63.08}} \\

\midrule

\multirow{3}*{RE} 
& NYT 		& 93.50 & - & - 
& 90.47 & \underline{\textbf{91.82}} & 50.42 
& 8.34 & 13.31 & 21.75 & \uuline{\textbf{43.37}} \\ 

& TACREV 	& 85.80 & $^\triangleright$59.40 | \quad-\quad~~ &  47.12 | \quad-~~~~
&  3.48 & 35.44 & \underline{\textbf{60.33}} 
& 16.67 & 19.80 & 46.16 & \uuline{\textbf{66.46}} \\ 

& FewRel 	& - & \quad-\quad | 84.20 & \quad-\quad | 96.51 
& 39.55 & 68.76 & \underline{\textbf{80.00}} 
& 45.60 & 42.53 & 61.80 & \uuline{\textbf{78.80}} \\

\midrule

\multirow{3}*{ED} 
& ACE05 	& 83.65 & 51.20 | 54.50 & \uuline{\textbf{55.61}} | 64.80 
& \underline{\textbf{77.13}} & 51.99 & 49.50 
& 13.81 & 20.63 & 38.75 & 53.94 \\ 

& MAVEN  	& 79.09 & $^\ast$\underline{\textbf{59.90}} | 36.86$^\diamond$ & ~\uuline{\textbf{64.80}} | 93.06$^\dagger$ 
& -     & 33.19 & 38.53 
& 7.22  & 9.49  & 25.29 & 33.39 \\ 

& FewEvent 	& 96.58 & \underline{\textbf{58.14}} | 68.37 & $^\ddagger$\uuline{\textbf{60.67}} | 93.18 ~
& 28.80 & 43.27 & 42.80 
& 21.82 & 25.23 & 35.36 & 46.44 \\ 

\midrule

\multirow{3}*{EAE} 
& ACE05 		& 73.50 & $^\triangleright$31.20 | 31.40 ~ & 45.90 | 42.70 
& \underline{\textbf{72.94}} & 35.62 & 23.60 / 47.35 
& 4.75 / 24.89  & 9.05 / 30.96 & 25.17 / 52.79 & 39.82 / \uuline{\textbf{58.54}} \\ 

& RAMS  		& 59.66 & - &  \uuline{\textbf{54.08}} | \quad-\quad\quad 
& -     &  6.08 & \underline{\textbf{49.08}} 
& 26.78 & 33.55 & 44.16 & 51.91 \\ 

& WikiEvents 	& 70.08 & - & - 
& 7.58  & 5.10  & \underline{\textbf{51.28}} 
& 20.00 & 21.37 & 38.20 & \uuline{\textbf{45.46}} \\


\bottomrule

\end{tabular}

}
\caption{Micro F1 results (marking the \textbf{best} all-way \underline{\textbf{0-shot}} \& \uuline{\textbf{5-shot}}) of previous SOTA and LLM-based methods. 
$\triangleright$: LLM-based; 
$\ast$: type-specific prompting \cite{ACL2023-Short_Prompt-EE_TypePrompt}; 
$\diamond$: prompt-based meta learning \cite{ACL2023_Meta-EE};
$\dagger$: 45-way-5-shot \cite{ACL2023_Meta-EE};
$\ddagger$: all-way-4-shot \cite{J2023_Prompt-EE_MsPrompt}; 
$r_o / r_w$ for ACE05 denotes the EAE result w/o or w/ golden triggers. 
Due to space limits, we list the methods achieving previous SOTA in Appendix~\ref{supp:pre_sota}.
\label{tab:exp} 
}

\end{table*}

\section{\fontsize{11.5pt}{0.1\baselineskip}\selectfont Empirical Study on Technical Solutions}
\label{sec:experiments}

\subsection{Setup}

\textbf{Datasets.} 
We adopt some widely used datasets for each subtask of low-resource {\ours}. 
\underline{\emph{NER}}: we use CoNLL03 \cite{CoNLL2003_CoNLL03}; OntoNotes5.0 \cite{Data2013_Ontonotes5.0}; FewNERD \cite{ACL2021_Few-NERD}. 
\underline{\emph{RE}}: we adopt TACREV \cite{ACL2020_TACREV}; NYT \cite{PKDD2010_NYT}; FewRel \cite{EMNLP2018_FewRel}.
\underline{\emph{ED}} of EE: we use ACE05 \cite{Book_ACE05}; MAVEN \cite{EMNLP2020_MAVEN}; FewEvent \cite{WSDM2020_MetaL-EE_DMBPN}. 
\underline{\emph{EAE}} (event argument extraction) of EE: we adopt ACE05 \cite{Book_ACE05}; RAMS \cite{DACL2020_RAMS}; WikiEvents \cite{NAACL2019_WikiEvents}. 

We follow \cite{EMNLP2023-Findings_LLM4LowResKE} to construct few-shot datasets in this paper.  
In all experiments, 
we utilize micro F1 score following previous methods like \cite{ACL2022_UIE} to evaluate performance.  
We release the code and data on GitHub\footnote{\url{https://github.com/mayubo2333/LLM_project}.}. 



\textbf{Models.} 
We conduct empirical study on LLM-based methods comparing with previous SOTA approaches. 
For prompt-/fine-tuning LLM, we adopt three typical models: 
\emph{InstructUIE} \cite{arXiv2023_PromptTuning_InstructUIE} with prompt-tuning on FlanT5-11B \cite{arXiv2022_FlanT5};  
\emph{KnowLM}\footnote{\url{https://github.com/zjunlp/KnowLM}.} which fine-tuned LLaMA \cite{MetaAI2023_LLaMA} with LoRA \cite{ICLR2022_LoRA}; 
and 
\emph{fine-tuning ChatGPT}\footnote{\url{https://platform.openai.com/docs/guides/fine-tuning}.}. 
For ICL with LLM, we select three widely used LLMs: 
\emph{Vicuna(1.5)}\footnote{\url{https://github.com/lm-sys/FastChat}.}; 
\emph{ChatGPT}\footnotemark[3]; 
\emph{GPT-4} \cite{OpenAI2023_GPT4}. 

\subsection{Implementation Details}

We test \emph{InstructUIE} and \emph{KnowLM} following their guidelines. 
In addition to the default parameters, 
for \underline{\emph{InstructUIE}}, we respectively set the maximum length for source, target and generation to 512, 50, 50, and set maximum instance number for each task to 200; 
for \underline{\emph{KnowLM}}, we use the version of \texttt{knowlm-13b-base-v1.0}, and respectively set the maximum length for source and new tokens to 512 and 300. 
For \underline{\emph{Vicuna}}, we adopt Vicuna-13B and Vicuna-1.5-13B without fine-tuning, setting the maximum input length to 1800; the batch size to 1; and both \texttt{frequency\_penalty} and \texttt{presence\_penalty} to 0. About the number of demonstrations, we set it to 4 for NER and ED; 8 for EAE; and 16 for RE. 
We run each experiment on 4 NVIDIA V100 GPUs for InstructUIE and KnowLM, and a single one for Vicuna. To save memory, we leverage the Accelerate framework and fp16 inference. 
For \underline{\emph{Fine-Tuning ChatGPT}}, we fine-tune \texttt{gpt-turbo-3.5-0613} on each specific dataset. We split 10\% samples from training set to construct valid set and train over 5 epochs with 5-shot samples, then conduct instruction prompting on fine-tuned ChatGPT. 
For \underline{\emph{ChatGPT}} and \underline{\emph{GPT-4}}, 
we call official APIs (\texttt{gpt-3.5-turbo-0301}/\texttt{gpt-4-0314}) without fine-tuning, and set the maximum input length to 3600 for all tasks. 
For \emph{ICL with LLMs (Vicuna(1.5), ChatGPT, GPT-4)}, 
We generate output with sampling temperature as 0 (\ie, greedy decoding).
We unify the maximum output length to 32 for RE; 96 for NER, ED and EAE. 

\subsection{Experimental Analysis}
\label{sec:exp_analysis}
In this section, we intend to analyze the performance of various (especially LLM-based) models, and explore the optimal selection of low-resource {\ours} approaches. 
We summarize main experimental results in Table~\ref{tab:exp}, and elaborate on our key findings as below: 

\noindent \textbf{Open-source or Proprietary LLMs?} 
We observe that prompt-/fine-tuning InstructUIE/KnowLM (\emph{open-source LLMs}) significantly outperform fine-tuned ChatGPT (\emph{proprietary LLMs}) on most tasks, except for under-trained datasets, \eg, TACREV, MAVEN, RAMS and WikiEvents. Nevertheless, ICL with Vicuna(1.5) (\emph{open-source LLMs}) lag far behind ChatGPT and GPT-4 (\emph{proprietary LLMs}).  
Thus we conclude that {\color{finding}\emph{for ICL, proprietary LLMs outperform open-source LLMs; when it comes to prompt-tuning or fine-tuning, open-source LLMs outperform their proprietary counterparts.}} 
We speculate that the possible reasons are: 
(1) The fundamental abilities of proprietary LLMs are much better (larger parameters, more abundant training data and effective training strategies), so that they understand the task from in-context demonstrations more effectively; 
(2) Open-source LLMs are much more lightweight than proprietary LLMs, contributing to thoroughly training and adapting to a specific task. While for proprietary LLMs like ChatGPT, the limited flexibility of fine-tuning APIs makes it difficult to obtain a well-trained task-specific model. 

\noindent \textbf{Tuning or ICL?} 
Regarding tuning InstructUIE and KnowLM, we find they achieve satisfying  performance on well-trained tasks (\eg, NER and RE) but show limited generalization abilities on under-trained tasks (\eg, ED and EAE). 
For ChatGPT, we observe that fine-tuning leads to generally consistent improvement on RE, ED and EAE tasks than ICL, especially on datasets which ICL with ChatGPT struggles, \eg, NYT, TACREV, ACE05 and MAVEN. However, even fine-tuned ChatGPT only shows similar performance with GPT-4 and performs worse than well-tuned SOTA small PLMs and open-resource LLMs. 
The possible reason is that for ICL with not-fine-tuned LLMs, where the instructions of LLMs are under alignment with the specific {\ours} task \cite{ACL2023-Findings_PromptTuning_RE}, it could be difficult to fully leverage the power of LLMs. 
As prompt-/fine-tuning enables adapting the LLM to specific tasks by adjusting its inputs or training it on specific data, we can infer that {\color{finding}\emph{tuning performs generally best in low-resource {\ours}, and more thoroughly training contributes to better performance}.}   

\noindent \textbf{Simple v.s. Complex Tasks.} 
Across different {\ours} tasks, we observe that LLMs are more proficient on NER and EAE (event class given) tasks than ED task. Given that ED is a more complex task (which requires instantiating abstract event classes on words in the context), we speculate that prompt-tuned LLMs are not well-acquainted with this task, and unable to fully understand the complex task through instructions and demonstrations. 
Furthermore, we find that the performance \textit{gaps} between small SOTA PLMs and LLMs are larger on the heterogeneous dataset (\eg, NYT, MAVEN, WikiEvents), suggesting that LLMs have difficulty in understanding the task containing numerous labels. 
The possible reason is related to low relevance between fine-grained labels and limited prompting input. 
Thus we infer that {\color{finding}\emph{LLMs struggle on more complex and fine-grained tasks.}}

\section{Benchmarks and Applications}
\label{sec:datasets_app}

\subsection{Benchmarks}

\begin{table}[!t] 
\centering
\small


\resizebox{\linewidth}{!}{

\begin{tabular}{c | c | c}

\toprule

\textbf{Task} & \textbf{Dataset} & \textbf{Summary} \\

\midrule

\tabincell{c}{ \textbf{Low-res} \\ \textbf{NER} } &
\tabincell{c}{ \texttt{Few-NERD} \\ \cite{ACL2021_Few-NERD} } & 
\tabincell{c}{ human-annotated; \\ 8 super- and 66 sub- entity types; \\ 188,238 sentences from Wikipedia \\ with 4,601,160 words. } \\ 

\midrule

\multirow{9}*{ \tabincell{c}{ \textbf{Low-res} \\ \textbf{RE} } } & 
\tabincell{c}{ \texttt{FewRel} \\ \cite{EMNLP2018_FewRel} } & 
\tabincell{c}{ human-annotated; \\ 100 relations;  70,000 sentences \\ derived from Wikipedia with \\ 124,577 unique tokens in total. } \\ 
\cmidrule{2-3} &
\tabincell{c}{ \texttt{FewRel2.0} \\ \cite{EMNLP2019_FewRel2} } & 
\tabincell{c}{ more challenging than FewRel, \\ considering biomedical domain \\ and NOTA setting. } \\ 
\cmidrule{2-3} &
\tabincell{c}{ \texttt{Entail-RE} \\ \cite{KBS2022_kHPN} } & 
\tabincell{c}{ human-annotated; \\ 80 relation types; 220k instances  from \\ Wikipedia; 19 relation entailment pairs. } \\

\midrule

\multirow{9}*{ \tabincell{c}{ \textbf{Low-res} \\ \textbf{EE} } } & 
\tabincell{c}{ \texttt{FewEvent} \\ \cite{WSDM2020_MetaL-EE_DMBPN} } & 
\tabincell{c}{ human- \& machine- annotated; \\ 19 super- and 100 sub- event types; \\ 70,852 instances from Wikipedia. } \\ 
\cmidrule{2-3} &
\tabincell{c}{ \texttt{Causal-EE} \\ \cite{KBS2022_kHPN} } & 
\tabincell{c}{ human-annotated; \\ 80 event types; 3.7k instances from \\ Wikipedia; 112 event causality pairs. } \\ 
\cmidrule{2-3} &
\tabincell{c}{ \texttt{OntoEvent} \\ \cite{ACL2021_OntoED} } & 
\tabincell{c}{ human- \& machine- annotated; \\ 13 super- and 100 sub- event types; \\ 60,546 instances from Wikipedia; \\ 3,804 event correlation pairs. } \\

\bottomrule

\end{tabular}

}


\caption{
A summary of some publicly-released low-resource {\ours} benchmarks.
\label{tab:datasets} 
}


\end{table}

In this section, we briefly introduce some public benchmarks specially established for low-resource {\ours}, as shown in Table~\ref{tab:datasets}. 
Besides, there are also some low-resource {\ours} methods \cite{EMNLP2020_RCEE} directly sampling part of data from public {\ours} benchmarks in general scenarios. 


\textbf{Low-resource NER.} 
\texttt{Few-NERD} \cite{ACL2021_Few-NERD} is designed for \emph{few-shot NER}, 
where each fine-grained entity type has sufficient examples for few-shot learning, 
including \texttt{Few-NERD (INTRA)} and \texttt{Few-NERD (INTER)}. 
All entities in different sets of \texttt{Few-NERD (INTRA)} belong to different types, in contrast with \texttt{Few-NERD (INTER)}. 

\textbf{Low-resource RE.}
\texttt{FewRel} \cite{EMNLP2018_FewRel}, designed for \emph{few-shot RE}, is a large-scale supervised dataset. 
\texttt{FewRel2.0} \cite{EMNLP2019_FewRel2} is derived from \texttt{FewRel}, and used for more challenging tasks: (1) adapt to a new domain with only a handful of instances, and (2) detect none-of-the-above (NOTA) relations. 
\texttt{Entail-RE} \cite{KBS2022_kHPN}, designed for \emph{low-resource RE}, is augmented with relation entailment annotations. 
\texttt{Wiki-ZSL} \cite{NAACL2021_RE-ZSBERT} is proposed for \emph{zero-shot RE}. 
More low-resource RE studies are shown in FewREBench \cite{EMNLP2022_FewREBench}.

\textbf{Low-resource EE.}
\texttt{FewEvent} \cite{WSDM2020_MetaL-EE_DMBPN} designed for \emph{few-shot ED}, is derived from widely-used EE datasets and data augmentation. 
\texttt{Causal-EE} \cite{KBS2022_kHPN} and \texttt{OntoEvent} \cite{ACL2021_OntoED}, designed for \emph{low-resource ED}, is augmented with event-relations. 
\texttt{FewDocAE} \cite{ACL2023_FewDocAE} is used for \emph{few-shot
document-level EAE}. 
More EE studies are shown in \textsc{TextEE} \cite{arXiv2023_ReEval4EE}.




\subsection{Applications}

We propose some promising applications of low-resource {\ours}, and summarize them into two aspects. 

(1) \emph{Domain-specific Applications} \cite{J2018_IE-Application}. 
With the continuous emergence of knowledge from various domains, such as health care, natural science, and finance, it's challenging to annotate large-scale structured knowledge.  
Especially for specific domains involving data privacy, where sufficient labeled data is not available, low-resource {\ours} is naturally crucial. 

(2) \emph{Knowledge-intensive Tasks} \cite{TNNLS2022_KE-Application}. 
Low-resource {\ours} has the potential to enhance knowledge-intensive tasks such as KG-based QA, fact verification, text-to-structure generation, and commonsense reasoning. 
Intuitively, those tasks require background KG for reasoning; however, collecting high-quality KGs is cumbersome and time-consuming. 
Low-resource {\ours} can efficiently collect diverse knowledge across resources and simultaneously enhance performance.  

\section{Comparison, Discussion and Outlook} 
\label{sec:compare_discuss}

\textbf{Comparison of Traditional Methods.}
From the methodology perspective, 
(1) \emph{Exploiting higher-resource data}, \emph{essentially to enrich the data features in low-resource scenarios}, serves as a general solution, which can be applied to widespread backbone models. 
(2) \emph{Developing stronger data-efficient models}, \emph{essentially to constrain the hypothesis space more precisely}, is a fundamental advancement to enable models equipped with specific low-resource learning abilities, typically through leveraging or transferring ``modeledge'' (\eg, implicit knowledge captured in model weights) in PLMs.  
(3) \emph{Optimizing data \& models together}, \emph{essentially to pursue the best strategy for training on representative data and searching the optimal hypothesis space}, represents a synthesis of the aforementioned two paradigms. 

From the task perspective, low-resource NER and RE are two typical tasks that have been well investigated. 
Since their schema of NER and RE are relatively simple, they can normally obtain satisfactory performance, especially with the utilization of traditional PLMs and recent LLMs. 
However, low-resource EE still suffers from poor generalization due to the complicated task schema. Thus, it remains challenging but promising to resolve the EE issues of event interpretation and representation in low-resource scenarios.

\noindent
\textbf{Discussion on LLM-Based Methods.} 
Comparing to traditional PLMs, LLMs equipped with more powerful pretrained abilities enable more complex prompt learning, and can be given more complicated instructions. Similar to traditional PLMs, LLMs with limited input capacity also struggle to tackle {\ours} tasks with the intricate schema. 
Through the empirical study in $\S$\ref{sec:exp_analysis}, we deduce that \emph{tuning open-resource LLMs and ICL with GPT family is promising in general, and the optimal LLM-based technical solution for low-resource {\ours} can be task-dependent.} 

Furthermore, \cite{EMNLP2023-Findings_LLM4LowResKE} demonstrated that LLMs are generally not effective few-shot information extractors, but excel at reranking challenging samples. Besides, LLMs have potential to serve as data creators \cite{EMNLP2023_LLM-DataGen} for low-resource {\ours}. 
In addition, as LLMs allow for multi-agent interaction \cite{arXiv2023_MultiAgent,arXiv2023_MultiAgent_Social}, tool use \cite{EMNLP2023_Survey_ZS-NER,arXiv2023_ToolUse}, \etc, it's promising to achieve low-resource {\ours} through collaboration between different models, \eg, large and small LM cooperation. 

\textbf{Future Directions.} 
Despite many models were proposed as surveyed, potential directions remain:

\begin{itemize}

	\item \emph{Utilizing more informative knowledge} \cite{arXiv2023_LLM-KG_TingLiu,arXiv2023_LLM-KG_XindongWu,J2023_LLM-KG_Jeff,arXiv2023_LKM}. 
	Advanced sample selection techniques, like active learning, sample reranking \cite{EMNLP2023-Findings_LLM4LowResKE,EMNLP2023_DA-ReRank_NER} and data synthesizing \cite{EMNLP2023_LLM-DataGen,EMNLP2023-Findings_LLMAnnotator} with LLMs, are promising to efficiently identify and utilize informative samples, reducing large-scale annotation requirements. 
	
	\item \emph{Focusing on practical low-resource settings and applications.} 
	Current low-resource {\ours} models are mostly research-oriented which is unrealistic in real world settings. It is essential to explore practical low-resource {\ours} and applications, especially involving out-of-distribution (OOD) data, establishing benchmarks and evaluation metrics (\eg, low-computation/memory cost).

	\item \emph{Equipping models with adaptable inference.} 
	We assume that low-resource {\ours} should be robust to domain shifts, rather than being restricted to a specific domain. Thus, exploring lifelong low-resource {\ours} and OpenIE \cite{EMNLP2023_Eval-OpenIE,arXiv2023_LLM4OpenIE,arXiv2023_LLM4OpenKGC} then schema induction \cite{ACL2023-Demo_SchemaInduction} are promising. 
	
    \item \emph{Exploring robust, faithful and interpretable {\ours}.} 
    To help LMs exactly understand {\ours} targets and rationales \cite{ACL2023_Survey_RE} can also be effective, especially with utilization of LLMs. Additionally, incorporating human feedback in training can further enhance accuracy, fairness, and reduce risks of biased or discriminatory {\ours}.



\end{itemize}

\section{Conclusion}
\label{sec:cons}

In this paper, we present a literature review on low-resource {\ours} methodologies from \emph{traditional} and \emph{LLM-based} perspectives, systematically categorizing \textbf{traditional} methods into three paradigms: 
(1) exploiting \emph{higher}-resource data, 
(2) developing \emph{stronger} data-efficient models, and 
(3) optimizing data and models \emph{together}; 
and \textbf{LLM-based} methods into two paradigms: 
(1) direct inference without tuning, and 
(2) model specialization with tuning. 
We also summarize widely used benchmarks and suggest some applications. 
Furthermore, we compare traditional paradigms, discuss low-resource {\ours} with LLMs, and provide some insights into future work. 
Our survey aims to assist researchers in understanding this field and inspire innovative algorithms, while guiding industry practitioners in selecting appropriate technical solutions for real-world applications. 

\section*{Acknowledgment}
We would like to express gratitude to the anonymous reviewers for their kind and helpful comments. 
This work was supported by the National Natural Science Foundation of China (No. 62206246), the Fundamental Research Funds for the Central Universities (226-2023-00138), Zhejiang Provincial Natural Science Foundation of China (No. LGG22F030011), Yongjiang Talent Introduction Programme (2021A-156-G), Tencent AI Lab Rhino-Bird Focused Research Program (RBFR2024003), Information Technology Center and State Key Lab of CAD\&CG, Zhejiang University, and NUS-NCS Joint Laboratory (A-0008542-00-00).


\bibliography{custom}
\bibliographystyle{IEEEtran}

\section*{Appendices}

\appendix

\section{Previous SOTA Methods}
\label{supp:pre_sota}
For \textbf{Previous SOTA} results in Table~\ref{tab:exp}, we list the corresponding methods in Table~\protect\ref{tab:exp_pre_sota}. 

\begin{table*}[!htbp] 
\centering
\small



\begin{tabular}{c | l | l l l l l}

\toprule

\multirow{2}*{\textbf{Task}} & 
\multirow{2}*{\textbf{Dataset}} & 
\multicolumn{5}{c}{\textbf{Previous SOTA}} 
\\ 
& & full 
& all-way-0-shot & 10-way-0-shot  
& all-way-5-shot & 10-way-5-shot 
\\

\midrule

\multirow{3}*{NER} 
& CoNLL03 		& 94.60 \cite{ACL2021_NER-CoNLL03-full_SOTA} 
& 74.99 \cite{arXiv2023_ICL_NER}$^\triangleright$ & - 
& 83.25 \cite{AAAI2023_USM} & - 
\\

& OntoNotes5.0  & 92.30 \cite{J2023_NER-OntoNotes5.0-full_SOTA} 
& - & -
& 59.70 \cite{EMNLP2023-Findings_LLM4LowResKE}$^\star$ & -
\\ 

& FewNERD 		& 70.90 \cite{ACL2022_NER-FewNERD-full_SOTA}
& - & - 
& 59.41 \cite{EMNLP2023-Findings_LLM4LowResKE}$^\star$ & 79.00 \cite{EMNLP2023-Findings_Meta-NER}
\\

\midrule

\multirow{3}*{RE} 
& NYT 		& 93.50 \cite{IJCNN2022_RE-NYT-full_SOTA} 
& - & - 
& - & - 
\\ 

& TACREV 	& 85.80 \cite{arXiv2023_RE-TACREV-full_SOTA} 
& 59.40 \cite{ACL2023-Findings_PromptTuning_RE}$^\triangleright$ & - 
& 47.12 \cite{EMNLP2023-Findings_LLM4LowResKE}$^\circ$ & -
\\ 

& FewRel 	& - 
& - & 84.20 \cite{ACL2023-Findings_Prompt-RE}
& - & 96.51 \cite{J2023_Meta-RE}
\\

\midrule

\multirow{3}*{ED} 
& ACE05 	& 83.65 \cite{NAACL2022-Findings_Prompt-EE}
& 51.20 \cite{AAAI2023_USM} & 54.50 \cite{ACL2023-Short_Prompt-EE_TypePrompt}
& 55.61 \cite{NAACL2022_Prompt-EE} & 64.80 \cite{NAACL2022_Prompt-EE}
\\ 

& MAVEN  	& 79.09 \cite{ACL2023_SPEECH}
& 59.90 \cite{ACL2023-Short_Prompt-EE_TypePrompt}$^\ast$ & 36.86 \cite{ACL2023_Meta-EE}$^\diamond$
& 64.80 \cite{ACL2023_Survey_EE} & 93.06 \cite{ACL2023_Meta-EE}$^\dagger$
\\ 

& FewEvent 	& 96.58 \cite{NAACL2022-Findings_Prompt-EE}
& 58.14 \cite{NAACL2022-Findings_Prompt-EE} & 68.37 \cite{ACL2023_Meta-EE}
& 60.67 \cite{J2023_Prompt-EE_MsPrompt}$^\ddagger$ & 93.18 \cite{ACL2023_Meta-EE}
\\ 

\midrule

\multirow{3}*{EAE} 
& ACE05 	& 73.50 \cite{NAACL2022_Prompt-EE}
& 31.20 \cite{EACL2023-Findings_InstructionPrompting-EE}$^\triangleright$ & 31.40 \cite{NAACL2022_Prompt-EE}
& 45.90 \cite{EMNLP2023-Findings_LLM4LowResKE}$^\star$ & 42.70 \cite{NAACL2022_Prompt-EE}
\\ 

& RAMS  		& 59.66 \cite{ACL2023-Findings_EAE-others-full_SOTA}
& - & - 
& 54.08 \cite{EMNLP2023-Findings_LLM4LowResKE}$^\star$ & -
\\ 

& WikiEvents 	& 70.08 \cite{ACL2023-Findings_EAE-others-full_SOTA}
& - & - 
& - & -
\\

\bottomrule

\end{tabular}

\caption{
The methods achieving the previous SOTA micro F1 performance, based on results in Table~\ref{tab:exp}. 
$\triangleright$: LLM-based; 
$\ast$: type-specific prompting; 
$\diamond$: prompt-based meta learning;
$\dagger$: 45-way-5-shot;
$\ddagger$: all-way-4-shot. 
\label{tab:exp_pre_sota} 
}

\end{table*}

\end{document}